\documentclass[letterpaper, 10 pt, conference]{ieeeconf}

\IEEEoverridecommandlockouts                              

\usepackage{graphics} 
\usepackage{epsfig} 
\usepackage{mathptmx} 
\usepackage{times} 
\usepackage{amsmath} 
\usepackage{amssymb}  
\usepackage{upgreek}
\usepackage{calrsfs}
\DeclareMathAlphabet{\pazocal}{OMS}{zplm}{m}{n}
\usepackage{dsfont}
\usepackage{graphicx}
\usepackage{tabu}
\usepackage{capt-of}
\usepackage{graphicx}
\usepackage{caption}
\usepackage{subcaption}
\usepackage{siunitx}
\usepackage{balance}
\usepackage[font=scriptsize,labelfont=bf]{caption}
\usepackage{comment}
\usepackage{tabularx}
\usepackage{dblfloatfix}
\usepackage{url}
\usepackage{physics}
\usepackage{cite}

\DeclareMathOperator*{\argmax}{arg\,max}

\title{\LARGE \bf
3D Reconstruction in Noisy Agricultural Environments: \\
A Bayesian Optimization Perspective for View Planning 
}
\author{Athanasios Bacharis$^1$, Konstantinos D. Polyzos$^2$, Henry J. Nelson$^1$, \\ Georgios B. Giannakis$^2$ and Nikolaos Papanikolopoulos$^1$\\
\footnotesize{$\{$bacha035, polyz003, nels8279, georgios, papan001$\}$@umn.edu} \\
\it \normalsize $^1$Department of Computer Science and Engineering, University of Minnesota \\
\it \normalsize $^2$Department of Electrical and Computer Engineering, University of Minnesota \\
}

\begin{document}

\maketitle

\begin{abstract}
3D reconstruction is a fundamental task in robotics that gained attention due to its major impact in a wide variety of practical settings, including agriculture, underwater, and urban environments. This task can be carried out via view planning (VP), which aims to optimally place a certain number of cameras in positions that maximize the visual information, improving the resulting 3D reconstruction. Nonetheless, in most real-world settings, existing environmental noise can significantly affect the performance of 3D reconstruction. To that end, this work advocates a novel geometric-based reconstruction quality function for VP, that accounts for the existing noise of the environment, without requiring its closed-form expression. With no analytic expression of the objective function, this work puts forth an adaptive Bayesian optimization algorithm for accurate 3D reconstruction in the presence of noise. Numerical tests on noisy agricultural environments showcase the merits of the proposed approach for 3D reconstruction with even a small number of available cameras.
\end{abstract}

\section{Introduction}

Acquiring visual information is a key component in robotics for scene understanding, planning, and decision-making. In particular, the acquisition of 3D information, known as 3D scene reconstruction, has gained popularity in different robotic settings, including agricultural \cite{bacharis2022view, nelson2023robust,peng2017view, roy2017active }, underwater \cite{vidal2020multisensor}, and urban \cite{jing2016view, smith2018aerial} environments. In the agricultural domain, informative 3D reconstructions have a major environmental and financial impact, including limited water usage for sustainable farming practices and increased revenue from large crop cultivation, just to list a few \cite{zermas2018extracting}. Obtaining an informative 3D scene reconstruction is typically modeled as an optimization problem, termed view planning (VP) \cite{tarabanis1995survey, tarbox1995planning}, that aims to decide the optimal placement of a set of available cameras in the 3D space. This optimal placement will result in the acquisition of the necessary information and, hence, the effective reconstruction of an area of interest.
VP can be carried out offline, given a fixed number of cameras and a static model of the environment. The use of a VP approach can significantly improve the quality of the 3D reconstruction (typically expressed as point clouds) compared to a set of arbitrarily placed cameras, which usually results in low-density point clouds.


Nonetheless, obtaining the optimal camera placement can become a 
challenging optimization problem when the size of the environment becomes large; see e.g., urban environments. To cope with this challenge, existing methods have formulated VP as a discrete optimization problem to reduce the space of feasible solutions \cite{bacharis2022view, peng2019adaptive}. A class of methods employed to solve the VP problem is based on search algorithms; e.g., \cite{bacharis2022view,jing2016view}. Recently, neural networks have been utilized in the VP setting to get an estimate of the function that maps any camera placement to its corresponding reconstruction quality \cite{pan2022scvp}. 

\begin{figure}[t]
\centering
\includegraphics[width=0.48\textwidth]{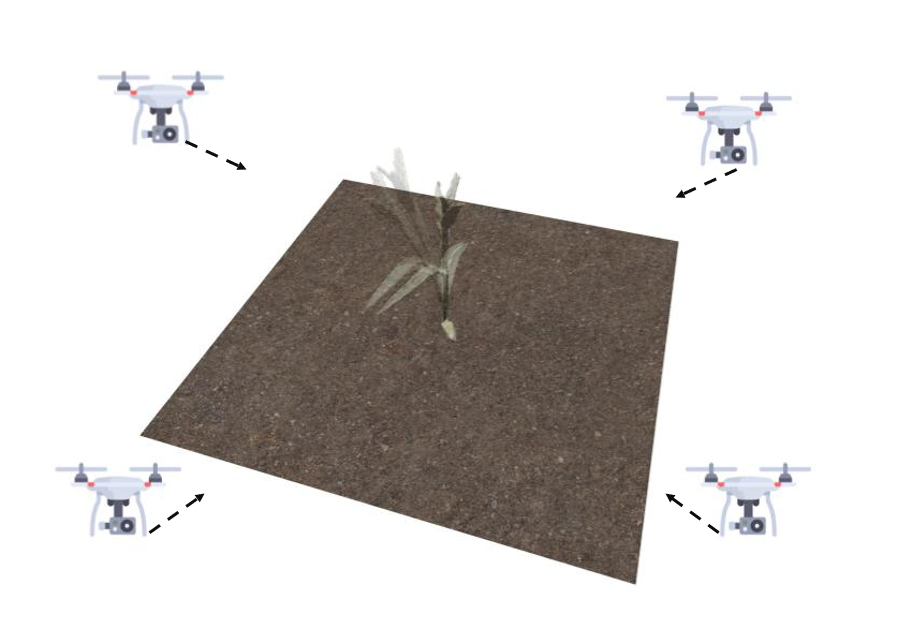}
\caption{Example of a noisy environment with four drones aiming to carry out the 3D reconstruction task (drone image \protect\footnotemark).}
\label{fig:teaser}
\end{figure}

\footnotetext{\url{https://www.pngwing.com/en/free-png-ddgwj/download}}






Albeit interesting, all these approaches do not account for any type of environmental noise, which can prove to be important when modeling the environment and can significantly affect the VP performance in practical settings. In Fig. \ref{fig:teaser}, there is an example of how the noise affects the geometry of the environment in VP. However, obtaining an analytic expression of the noise may be a challenging task, particularly when the number of noise realizations is not sufficient. With no analytic expression of the noise, the expression of the reward function that incorporates the noise also becomes \textit{unknown}, and thus conventional optimization techniques including gradient-based solvers, may not be applicable. 

 
This motivates the use of the Bayesian optimization (BO) paradigm that aims at optimizing a black-box (unknown) function by leveraging a (typically small) number of function evaluations to form a probabilistic \textit{surrogate model} that will guide the selection of new query points \cite{shahriari2015taking}. Relying on the surrogate model, the so-termed \textit{acquisition function} (AF), typically expressed in closed-form, is used to sequentially select the next query points in the optimization process. BO has been adopted in a range of practical settings including hyperparameter tuning  \cite{snoek2012practical}, drug discovery \cite{korovina2020chembo} and robotic-based tasks \cite{wang2017max,cully2015robots,marchant2014bayesian} to name a few. Nonetheless, none of the existing works has considered adopting BO in the VP problem under \textit{noisy} environments.

This paper presents a novel approach for modeling and optimizing the reward function of the VP problem in order to effectively and efficiently reconstruct an area of interest in the presence of noise. The contributions of this work can be summarized in the following directions

\begin{itemize}
    \item To the best of our knowledge, this work is the first to incorporate existing noise in the reward function of a VP problem for 3D reconstruction.
    \item To optimize the reward function with \textit{unknown} closed-form expression, the present work is the first to introduce a well-motivated adaptive BO method for the VP problem.
    \item Compared to most existing approaches that require discretization of the search space in order to be tractable, this work focuses on a \textit{continuous} optimization problem that considers the entire search space.  
    \item Promising experimental results on three different corn-based agricultural environments demonstrate the effectiveness of the proposed approach for accurate 3D reconstruction with a \textit{limited} number of available cameras. 
\end{itemize}


%

\section{Related Work}\label{sec:relwork}

\subsection{View Planning}

VP aims at placing cameras in an optimal way that maximizes the gain of information about the environment, to successfully carry out the 3D reconstruction task \cite{tarabanis1995survey, tarbox1995planning}. In order to reduce the complexity of the search space, which may be intractable in large environments, several works have modeled the problem in a discrete manner \cite{bacharis2022view, pan2022scvp, tarabanis1995survey, tarbox1995planning}. However, this discrete formulation is known to be NP-complete, as shown in \cite{tarbox1995planning}, leading to the use of sampling-based search methods \cite{bacharis2022view} or approximate algorithms that require two or more optimization phases \cite{peng2019adaptive}.


Besides an affordable complexity, a critical component in the VP optimization problem is the proper selection of the reward function. Given the information about the environment, two different types of reward functions have been mainly explored, namely occupancy maps and geometric evaluations. Although offering a simple representation of the environment, occupancy maps may reduce the quality of the 3D reconstruction \cite{vasquez2014view, pan2022scvp, pan2023one}. On the other hand, geometric functions typically achieve more dense point clouds in the resulting reconstruction by imposing more constraints in the view selection process \cite{bacharis2022view, peng2017view, roy2017active}. In our work, we will focus on a geometric function since we aim to improve the reliability of the 3D reconstruction.

Finally, an aspect of great interest in real-world settings is the existing noise of the environment. Although the information of the \textit{static} environment is given, in several applications there are many parameters, such as the wind, that may alter the representation of the environment, hence affecting the outcome of the 3D reconstruction. Therefore, it is of paramount importance to incorporate these types of disturbances in the selection of the views to achieve the best reconstruction.

Despite the effectiveness of the aforementioned VP approaches in noise-free environments, none of them has considered the effects of the noise in the VP process, or the merits of a continuous formulation over a discrete one where any position in the 3D space can be explored. 


\subsection{Bayesian Optimization}

Bayesian optimization (BO) provides a principled framework to optimize a \textit{black-box} or \textit{costly} to evaluate function, by judiciously exploiting a number of function values to build a surrogate model, that allows for the sequential acquisition of new points. The key components of BO are (i) the selection of a probabilistic surrogate model and (ii) the design of the AF to select new points to query on-the-fly. Gaussian processes (GPs) have been extensively adopted as a \textit{nonparametric} Bayesian surrogate model in a range of BO applications due to their well-documented merits in learning an unknown function along with its probability density function (pdf) in a sample-efficient manner \cite{Rasmussen2006gaussian, shahriari2015taking}. Although interesting and effective in several practical settings, their performance hinges on the proper selection of the kernel function to evaluate the pairwise similarity of different input points, whose selection is a non-trivial task. To cope with this challenge, existing works adaptively learn the kernel as new data arrive on-the-fly; see e.g., \cite{polyzluPAMI, polyzos2023bayesian, nguyen2020bayesian, gopakumar2018algorithmic, polyzos2022AL}. Capitalizing on GP-based surrogate models, several AFs have been designed including Thompson sampling \cite{thompson1933likelihood}, expected improvement (EI) \cite{jones1998efficient} or upper confidence bound \cite{srinivas2010gaussian} to name a few. These AFs are expressed in closed-form, they are easy to optimize and have well-documented merits in balancing \textit{exploration} and \textit{exploitation} of the search space. 

\section{Problem Formulation}\label{sec:probform}

The objective of the VP problem is to find an optimal camera setup that will maximize the information that can be extracted from a given environment in order to assist the 3D reconstruction task. Specifically, given a number of cameras $N$, and a point cloud $\{\mathbf{p}_i \}_{i=1}^{\pazocal{P}} \subseteq \mathbb{R}^3$, sampled from the given representation of the environment, the goal is to find

\begin{equation}
    \mathbf{z}^{*} =  \argmax_{\mathbf{z} \in \pazocal{Z}} \; r(\mathbf{z}; \mathbf{p}_1, \dots , \mathbf{p}_\pazocal{P})
    \label{eq:opt_r}
\end{equation}
where $\mathbf{z} := [ \mathbf{x}_1, \dots, \mathbf{x}_N ]$, with $\{ \mathbf{x}_i \}_{i=1}^N \subseteq \mathbb{R}^6$ denoting each camera feature vector and $\pazocal{Z}$ is the set of feasible solutions. In particular, $\mathbf{x}_i = [\mathbf{c}_i, \mathbf{v}_i]$ with $\mathbf{c}_i \in \mathbb{R}^3$ and $\mathbf{v}_i \in \mathbb{R}^3$ containing the information of the location, and the orientation of the $i^{\text{th}}$ camera. The reward function $r(\cdot)$ represents the quality of the 3D reconstruction.

Following the discrete geometrical formulation of \cite{bacharis2022view} for the $r(\cdot)$ function, an extended version of the $r(\cdot)$ in the continuous space is expressed as
\begin{equation}
    r(\mathbf{z}; \mathbf{p}_1, \dots , \mathbf{p}_\pazocal{P}) = \frac{1}{\pazocal{P}\pazocal{S} } \sum_{\substack{\mathbf{p}\in\pazocal{P}\\(\mathbf{c}_i, \mathbf{v}_i,\mathbf{c}_j,\mathbf{v}_j) \in \mathbf{z}}} f(\mathbf{c}_i,\mathbf{c}_j;\mathbf{p})  g(\mathbf{c}_i, \mathbf{v}_i,\mathbf{c}_j,\mathbf{v}_j;\mathbf{p})
    \label{eq:r}
\end{equation}
where $f$ indicates the reconstruction quality per point $\mathbf{p}$ from the camera location pair $(\mathbf{c}_i, \mathbf{c}_j)$, $g$ represents the geometric constraints for each $(\mathbf{c}_i,\mathbf{v}_i), (\mathbf{c}_j,\mathbf{v}_j)$ camera pair, and
$\pazocal{S}$ is the total number of $(i,j)$ pairs.

The geometric reconstruction quality, defined by $f$, relies on the angle between the vectors $(\mathbf{c}_i - \mathbf{p}),(\mathbf{c}_j-\mathbf{p})$ $\forall (i,j)$. When the angle value increases, the reconstruction quality error of $\mathbf{p}$ becomes smaller. This can be expressed by the sine of the angle between these two vectors as
\begin{equation}
    f(\mathbf{c}_i,\mathbf{c}_j;\mathbf{p}) = \frac{\norm{(\mathbf{c}_i - \mathbf{p}) \times (\mathbf{c}_j - \mathbf{p})}}{\norm{\mathbf{c}_i-\mathbf{p}}\norm{\mathbf{c}_j-\mathbf{p}}}
    \label{eq:f}
\end{equation}

In order for Eqn. \eqref{eq:f} to provide an accurate estimate of the reconstruction quality for each point $\mathbf{p}$, two conditions need to be satisfied. The first focuses on the number of those points that are able to be seen from each camera $(\mathbf{c}_i,\mathbf{v}_i)$, and the second on the ability of all the camera pairs $(\mathbf{c}_i,\mathbf{v}_i),(\mathbf{c}_j,\mathbf{v}_j)$ to gain 3D information of each common viewed point $\mathbf{p}$. Specifically, the function $g$ has the form
\begin{equation}
    g(\mathbf{c}_i, \mathbf{v}_i,\mathbf{c}_j,\mathbf{v}_j;\mathbf{p}) = \mathds{1}(cond_1^i) \mathds{1}(cond_1^j) \mathds{1}(cond_2^{i,j})
    \label{eq:g}
\end{equation}
where $\mathds{1}(\cdot)$ is the indicator function, returning value 1 when the input condition is satisfied and 0 otherwise. The conditions ($cond_1^i, cond_2^{i,j}$) regarding each camera pair ($i,j$) are defined as



\begin{equation}
    cond_1^i:  = \cos(\phi_1^i)- \cos (\frac{FoV}{2}) \geq 0
    \label{eq:cond1}
\end{equation}
\begin{equation}
    cond_2^{i,j}:  =  \cos(\phi_2^{i,j}) - \cos (\theta_{match}) \geq 0 
    \label{eq:cond2}
\end{equation}

where $ \cos(\phi_1^i): = \frac{(\mathbf{c_i}-\mathbf{p}) \cdot \mathbf{v_i}}{\norm{\mathbf{c_i}-\mathbf{p}} \norm{\mathbf{v_i}} }$, $ \cos(\phi_2^{i,j}): =\frac{(\mathbf{c}_i - \mathbf{p}) \cdot (\mathbf{c}_j - \mathbf{p})}{\norm{\mathbf{c}_i - \mathbf{p}} \norm{\mathbf{c}_j - \mathbf{p}} } $, $FoV$ is the camera field-of-view parameter and $\theta_{match}$ the maximum angle between the vectors ($\mathbf{c}_i - \mathbf{p}$), ($\mathbf{c}_j - \mathbf{p}$), so that to reconstruct the point $\mathbf{p}$.



The reward function in Eqn. \eqref{eq:opt_r}, although formulating the VP problem, does not account for real-world scenarios where there is noise in the environment. To alleviate this limitation, we assume that each point $\mathbf{p}$ under the noisy environment is modeled as $\tilde{\mathbf{p}} = h(\mathbf{p})$, where the function $h(\cdot)$ represents the effect of the noise, whose expression is considered unknown. For instance, the function $h(\cdot)$ may have a linear form of $h(\mathbf{p}) = \mathbf{p} + \mathbf{n}$, with $\mathbf{n}$ following a distribution $\mathcal{D}$ parameterized by $\omega$; that is $\mathbf{n} \sim \mathcal{D}(\omega)$. Taking into account the function $h(\cdot)$, the equivalent noisy reward function becomes
\begin{equation} 
    \begin{split}
        \tilde{r}(\mathbf{z}) = r(\mathbf{z}; \tilde{\mathbf{p}}_1, \dots , \tilde{\mathbf{p}}_\pazocal{P}).
    \end{split}
    \label{eq:r_tild}
\end{equation}
Finally, the VP optimization problem turns into
\begin{equation}
    \mathbf{z}^{*} =  \argmax_{\mathbf{z}\in \pazocal{Z}} \tilde{r}(\mathbf{z}).
    \label{eq:opt_r_noisy}
\end{equation}

It is worth noticing that the analytic expression of the objective function in Eqn. \eqref{eq:opt_r_noisy} becomes unknown when apriori information about the function $h(\cdot)$ is not provided. With no analytic expression at hand, conventional gradient-based solvers are not applicable for obtaining $\mathbf{z}^*$. This motivates well the BO paradigm that offers principled methods to effectively optimize black-box functions in a data-efficient manner, as outlined next.  

\begin{figure}[t]
\centering
\includegraphics[width=0.48\textwidth]{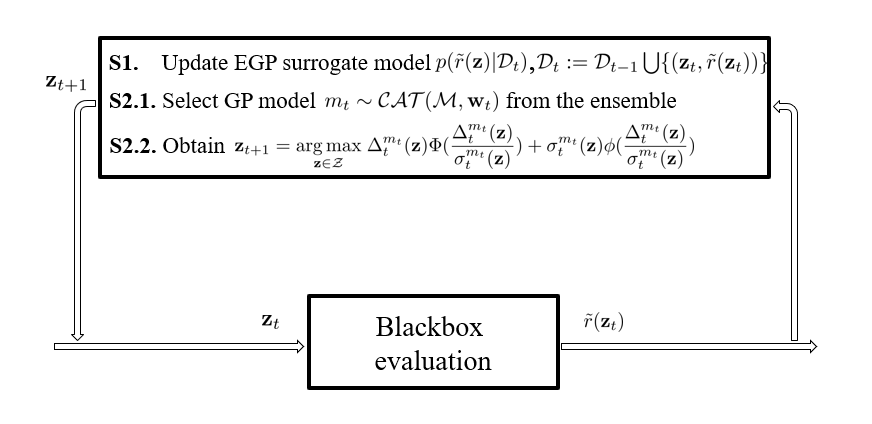}
\caption{EGP-VP in a nutshell.}
\label{fig:BO_nutshell}
\end{figure}

\begin{figure}[t]
\centering
\subfloat[\centering Case of 1-plant.]{\includegraphics[width=.24\textwidth]{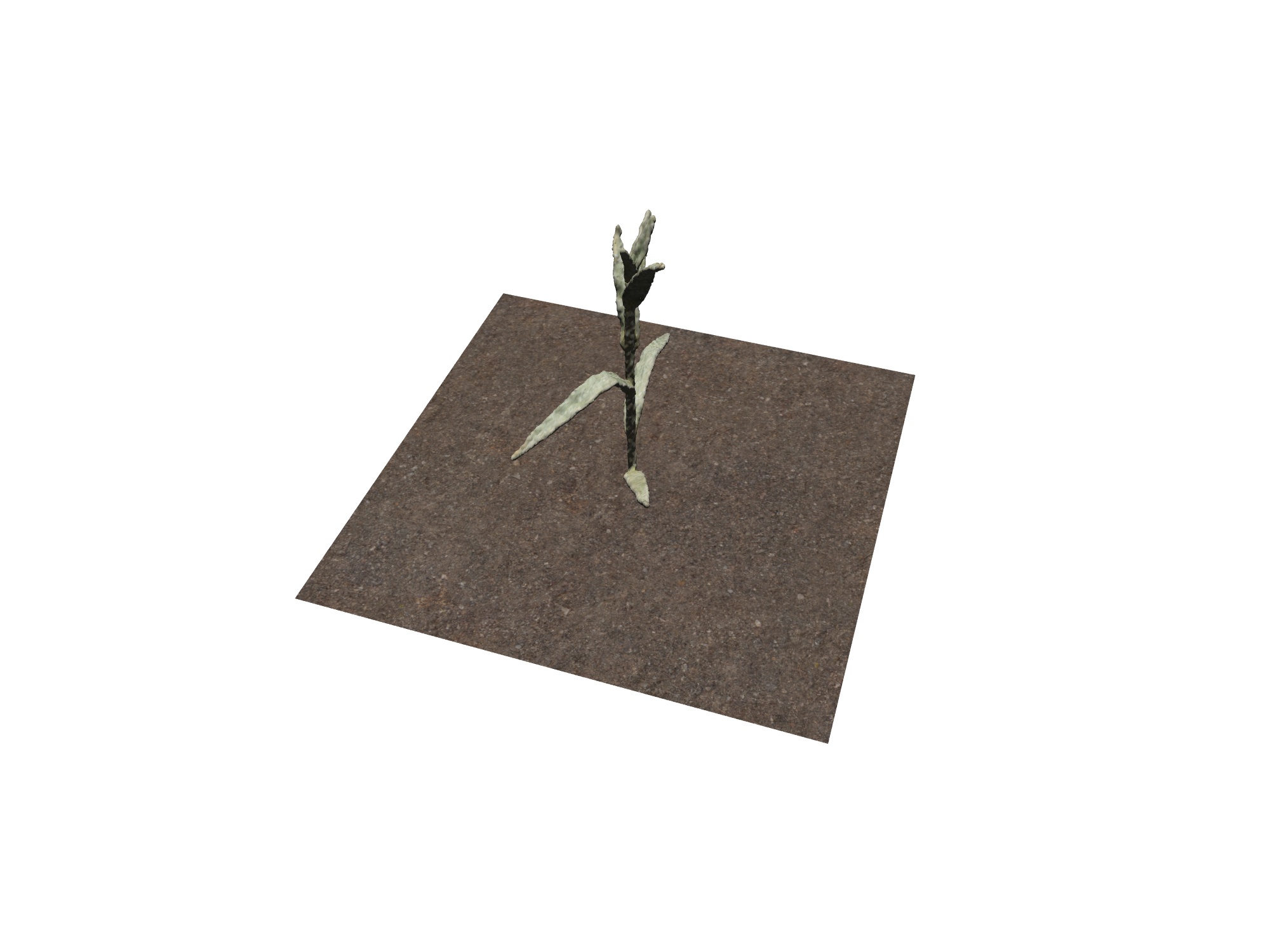}}\label{fig:1_plants}\hfill
\subfloat[\centering Case of 3-plant (row).]{\includegraphics[width=.24\textwidth]{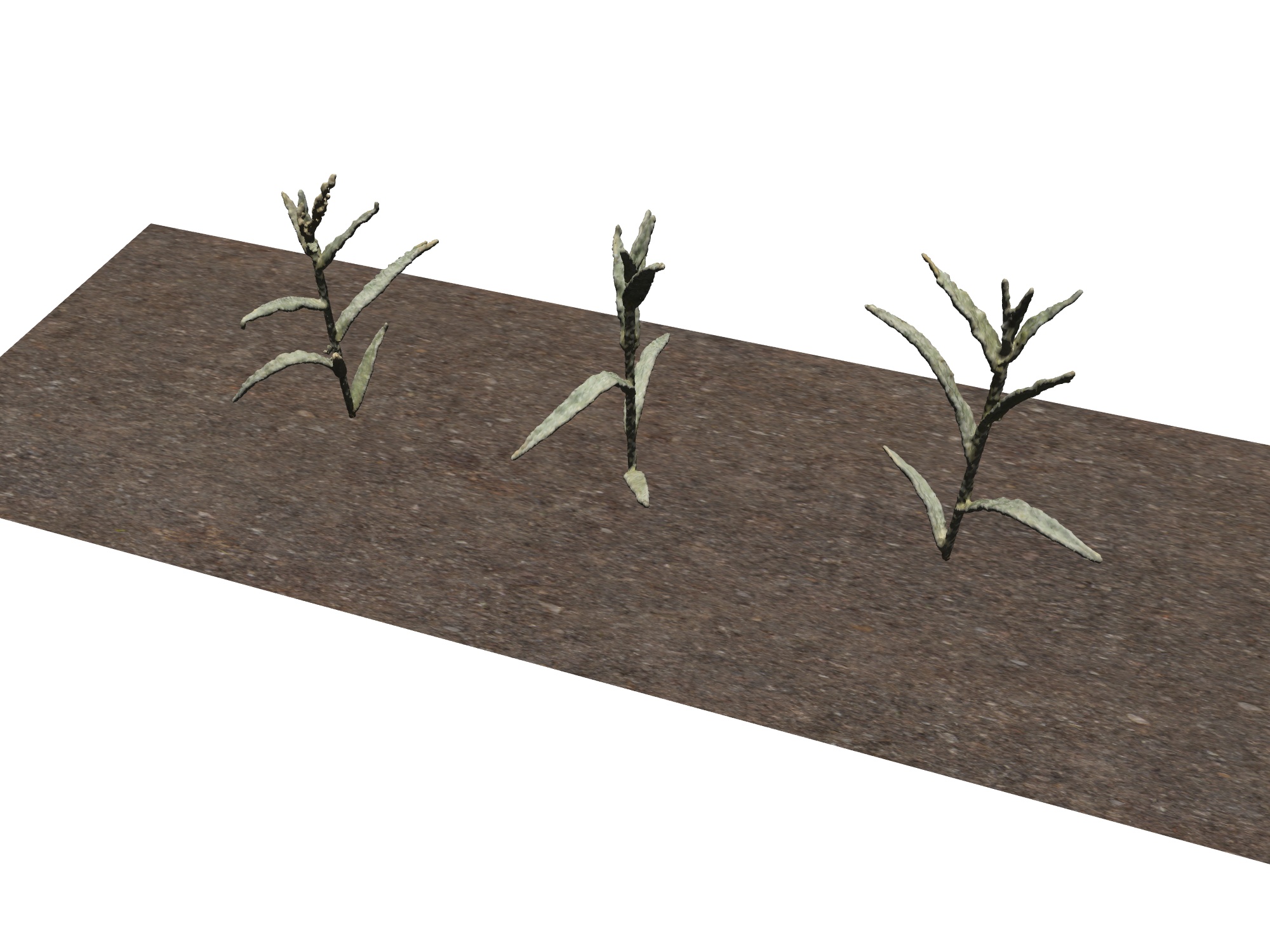}}\label{fig:3_plants}\hfill
\subfloat[\centering Case of 6-plant (rectangle).]{\includegraphics[width=.24\textwidth]{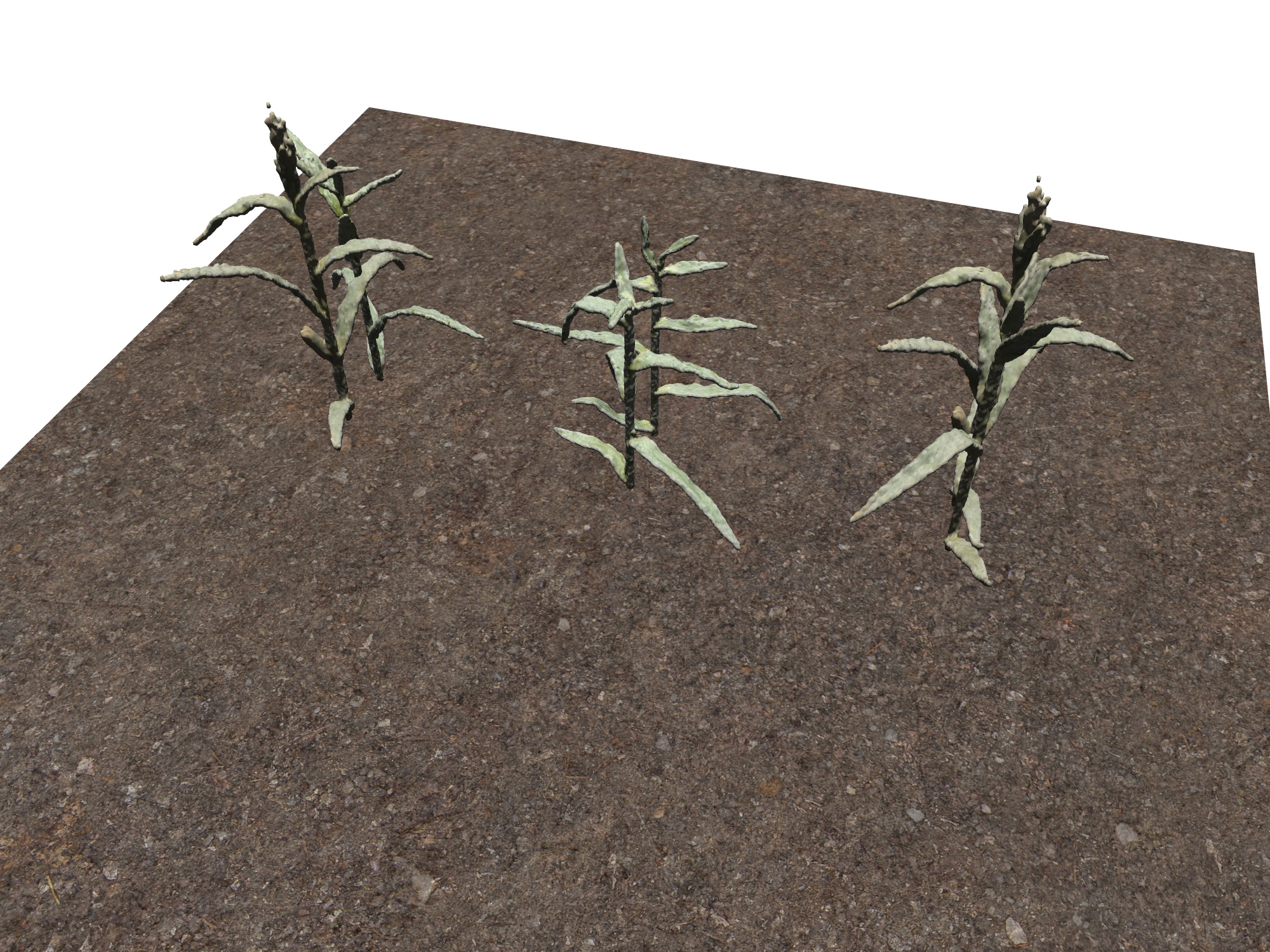}}\label{fig:6_plants}\hfill
\caption{Example of the three environmental cases.}
\label{fig:env_scenarios}
\end{figure}

\begin{figure*}[t]
\centering
\includegraphics[width=0.95\textwidth]{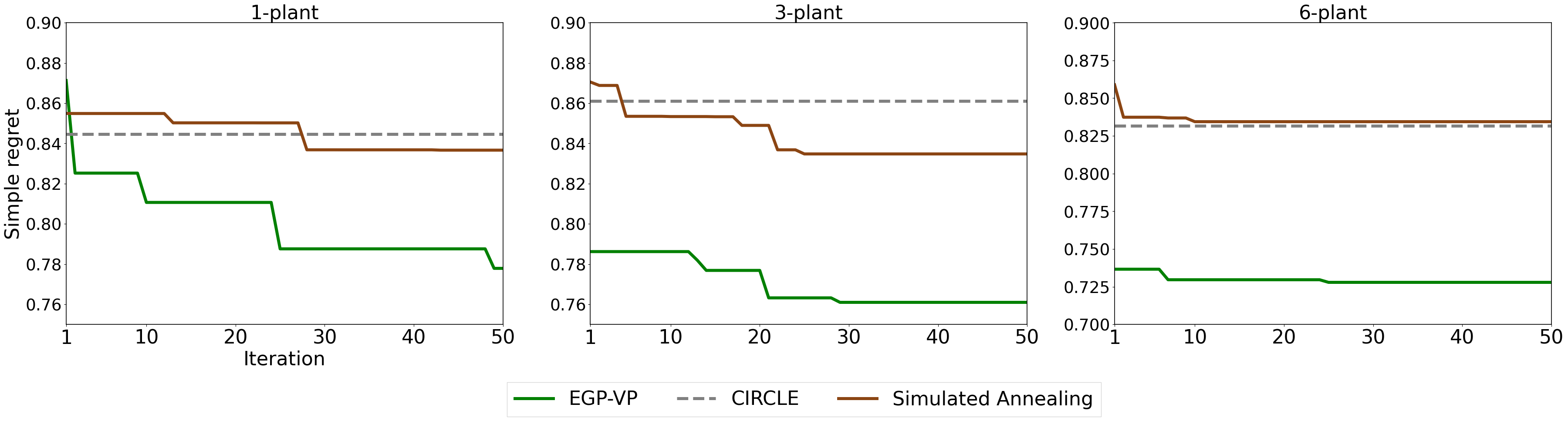}
\caption{Simple regret for (a) 1-plant, (b) 3-plant, and (c) 6-plant scenarios.}
\label{fig:sr}
\end{figure*}

\begin{figure*}[th]
     \centering
     \begin{subfigure}[t]{0.3\textwidth}
         \centering    \includegraphics[width=1.0\textwidth]{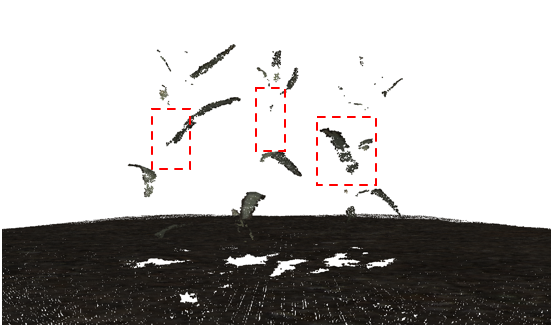}
         \vspace{-1em}
         \caption{Circle}
         \label{fig:c31}
     \end{subfigure}
     \begin{subfigure}[t]{0.3\textwidth}
         \centering  \includegraphics[width=1.0\textwidth]{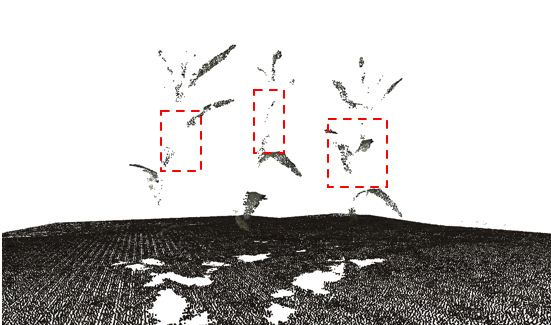}
         \vspace{-1em}
         \caption{Simulated Annealing}
         \label{fig:an31}
     \end{subfigure}
     \begin{subfigure}[t]{0.3\textwidth} \centering\includegraphics[width=1.0\textwidth]{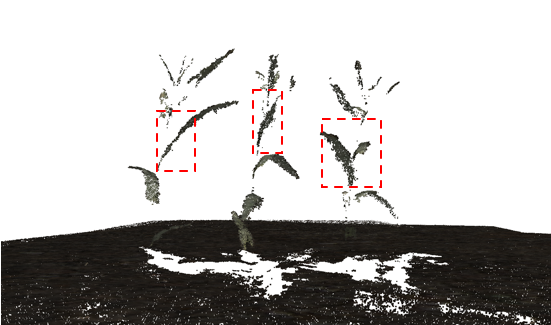}
        \vspace{-1em}
         \caption{EPG-VP}
         \label{fig:egp31}
     \end{subfigure}
     \caption{Reconstruction results from all competing methods shown on the 3-plant case from the same viewpoint to demonstrate differences.}
     \label{fig:recs}
\end{figure*}

\section{Optimization Method}\label{sec:optmethod}

Relying on a given budget of input-output data pairs $\pazocal{D}_t:=\{(\mathbf{z}_\tau, \tilde{r}_\tau)\}_{\tau=1}^t$ with $\tilde{r}_\tau:= \tilde{r}(\mathbf{z}_\tau)$, BO capitalizes on a probabilistic surrogate model $p(\tilde{r}(\mathbf{z})|\pazocal{D}_t)$ for $\tilde{r}$, that guides the acquisition of the next query input $\mathbf{z}_{t+1}$. Specifically, each iteration of the BO process follows a two-step approach; that is (i) update $p(\tilde{r}(\mathbf{z})|\pazocal{D}_t)$ using $\pazocal{D}_t$ and (ii) obtain $\mathbf{z}_{t+1}\! =\! \underset{\mathbf{z}\in\pazocal{Z}}{\arg\max} \ \alpha_{t+1} (\mathbf{z}|\pazocal{D}_t)$ using $p(\tilde{r}(\mathbf{z})|\pazocal{D}_t)$. The so-termed acquisition function (AF) $\alpha_{t+1} (\mathbf{z}|\pazocal{D}_t)$, often expressed in closed-form, is chosen so as to balance \textit{exploration} and \textit{exploitation} of the search space \cite{shahriari2015taking}. There are several choices for both $p(\tilde{r}(\mathbf{z})|\pazocal{D}_t)$ and $\alpha_{t+1} (\mathbf{z}|\pazocal{D}_t)$. In this work, we will focus on the Gaussian process (GP) based surrogate model that has been extensively used in a gamut of BO applications; see e.g.,  \cite{shahriari2015taking, polyzluPAMI}, and will use the expected improvement (EI) as AF.  
\vspace{-0.08cm} 
\subsection{BO with single GP and EI}\label{sec:sgp}

Belonging to the family of nonparametric Bayesian approaches, GPs offer a principled framework to learn an unknown nonlinear function with well-quantified uncertainty, which can readily guide the acquisition of new query input points. Learning with GPs begins with the assumption that a GP prior is postulated over the function $\tilde{r}(\cdot)$ as $\tilde{r} \sim \pazocal{GP}(0, \kappa(\mathbf{z},\mathbf{z}'))$, with $\kappa(\mathbf{z},\mathbf{z}')$ representing a positive-definite kernel function that measures the pairwise similarity between $\mathbf{z}$ and $\mathbf{z}'$. This means that the random vector $\tilde{\mathbf{r}}_t:= [\tilde{r}(\mathbf{z}_1),\ldots,\tilde{r}(\mathbf{z}_t)]^\top$ comprising the function values at inputs $\mathbf{Z}_t := \left[\mathbf{z}_1, \ldots, \mathbf{z}_t\right]^\top$ is Gaussian distributed as $\tilde{\mathbf{r}}_t \sim \pazocal{N} (\tilde{\mathbf{r}}_t ; {\bf 0}_t, {\bf K}_t) (\forall t)$, where ${\bf K}_t$ is the $t \times t$ kernel (covariance) matrix whose $(i,j)$th element is $[{\bf K}_t]_{i,j} = {\rm cov} (\tilde{r}(\mathbf{z}_i), \tilde{r}(\mathbf{z}_j)):=\kappa(\mathbf{z}_i, \mathbf{z}_j)$ \cite{Rasmussen2006gaussian}.    

With $\mathbf{y}_t := [y_1, \ldots, y_t]^\top$ denoting the (possibly) noisy output vector where $y_\tau = \tilde{r}({\bf z}_\tau) + n_\tau$ ($\forall \tau$) and $n_\tau \sim \pazocal{N}(n_\tau;0,\sigma_n^2)$ is white Gaussian noise uncorrelated across $\tau$, 
it can be shown that the function posterior pdf of $\tilde{r}(\mathbf{z})$ at any input $\mathbf{z}$ is Gaussian distributed as
\begin{align}
p(\tilde{r}(\mathbf{z})|\pazocal{D}_t) = \pazocal{N}(\tilde{r}(\mathbf{z}); \mu_t (\mathbf{z}), \sigma_{t}^2(\mathbf{z})) 
\end{align}
with mean and variance given in closed form as described in \cite{Rasmussen2006gaussian} 
\vspace{-0.1cm}
\begin{subequations}
\begin{align}	
\mu_t ({\bf z}) & = \mathbf{k}_t^{\top} ({\bf z}) (\mathbf{K}_t + \sigma_{n}^2
 \mathbf{I}_t)^{-1} \mathbf{y}_t \label{eq:mean}\\
\sigma_{t}^2 ({\bf z})& = \!\kappa(\mathbf{z},\mathbf{z})\! -\! \mathbf{k}_t^{\top} ({\bf z}) (\mathbf{K}_t\! +\! \sigma_{n}^2 \mathbf{I}_t)^{-1} \mathbf{k}_t ({\bf z}) \label{eq:variance}
\end{align}\label{eq:plain_gpp}
\end{subequations}
where $\mathbf{k}_t ({\bf z}) := [\kappa(\mathbf{z}_1, \mathbf{z}), \ldots, \kappa(\mathbf{z}_t,  \mathbf{z})]^\top$. 
Note that the mean in Eqn. \eqref{eq:mean} provides a point estimate of the function $\tilde{r}$ evaluated at $\mathbf{z}$ and the variance in Eqn. \eqref{eq:variance} quantifies the associated uncertainty.

Leveraging the posterior pdf $p(\tilde{r}(\mathbf{z})|\pazocal{D}_t)$ one can select the next input query point $\mathbf{z}_{t+1}$ using the EI AF as follows \cite{jones1998efficient}
\begin{align}
	\mathbf{z}_{t+1} = \underset{\mathbf{z}\in\pazocal{Z}}{\arg\max} \  \mathbb{E}_{p(\tilde{r}(\mathbf{z})|\pazocal{D}_t)}[\text{max}(0,\tilde{r}(\mathbf{z})-\hat{\tilde{r}}_t^{\text{max}})]   \label{eq:EI_AF}
\end{align}
with $\hat{\tilde{r}}_t^{\text{max}}$ denoting an estimate of the maximum value of $\tilde{r}(\cdot)$ at iteration $t$. A typical choice for this estimate is $\hat{\tilde{r}}_t^{\text{max}} = \text{max}(y_1,\dots,y_t)$; see e.g \cite{frazier2018tutorial, jones1998efficient}. When the GP surrogate model is adopted, the posterior pdf $p(\tilde{r}(\mathbf{z})|\pazocal{D}_t)$ is Gaussian distributed (c.f. Eqn. \eqref{eq:plain_gpp}) and hence the AF optimization task in Eqn. \eqref{eq:EI_AF} boils down to  
\begin{align}
	\mathbf{z}_{t+1} = \underset{\mathbf{z}\in\pazocal{Z}}{\arg\max} \   \Delta_t(\mathbf{z})\Phi\left(\frac{\Delta_t(\mathbf{z})}{\sigma_t(\mathbf{z})}\right)
    \!+\! \sigma_t(\mathbf{z})\phi\left(\frac{\Delta_t(\mathbf{z})}{\sigma_t(\mathbf{z})}\right) \label{eq:Gaussian_EI}  
\end{align}
where $\Delta_t(\mathbf{z}):= \mu_t(\mathbf{z})-\hat{\tilde{r}}_t^{\text{max}}$ and $\Phi(\cdot), \phi(\cdot)$ represent the Gaussian cumulative density function and Gaussian pdf respectively. The EI AF is one of the most prevalent acquisition criteria in BO due to its well-documented merits in balancing well exploration and exploitation of the search space \cite{jones1998efficient}. This is particularly appealing in the VP problem under noisy environments introduced in the present work, since the $\tilde{r}(\cdot)$ function is unknown with no prior information about the noise, and the goal is to find the function optimum relying solely on some available function values. 

\subsection{BO with ensemble GPs and adaptive EI}\label{sec:egpvp}

Although interesting and effective in several practical settings, the performance of the single GP-EI method hinges on the kernel function, whose apriori proper selection is a nontrivial task. To cope with this challenge, an ensemble of $M$ GP models (EGPs) is advocated to model $\tilde{r}(\cdot)$ which is \textit{agnostic} to the appropriate kernel at first, but will iteratively \textit{learn} the latter as new input-output data will become available. Each GP model $m \in \pazocal{M}:= \{1,\ldots, M \}$ relies on a distinct kernel function $\kappa^m(\cdot,\cdot)$ implying that for each model $m$, a unique GP prior is placed on $\tilde{r}$ as $\tilde{r}|m \sim \pazocal{GP}(0, \kappa^m (\mathbf{z}, \mathbf{z}'))$. Note that the kernel set $\pazocal{K}:=\{\kappa_m\}_{m=1}^M$ comprises kernels of different types and with different hyperparameters. Combining the $M$ GP priors in the ensemble with the initial weights $\{w_0^m\}_{m=1}^M$ forms the ensemble prior for $\tilde{r}$, which is a Gaussian mixture (GM) written as $\tilde{r}\sim \sum_{m=1}^M w^m_0 \pazocal{GP}(0,\kappa^m (\mathbf{z},\mathbf{z}'))$ with $\sum_{m=1}^M w^m_0 =1$. Each weight $w^m_0:={\rm Pr} (i=m)$ represents the probability that captures the influence of GP model $m$ in the ensemble. 

Capitalizing on the ensemble GM prior and the available budget of input-output pairs $\pazocal{D}_t$ at slot $t$, one can obtain the ensemble posterior pdf via the sum-product rule as 
\begin{align}
	{p}(\tilde{r}(\mathbf{z})|\pazocal{D}_{t}) \! &=\! \sum_{m = 1}^M\! w_t^m {p}(\tilde{r}(\mathbf{z})|  i\!=\! m,\pazocal{D}_{t} ) \label{eq:EGP_post}
\end{align}
which is a GM posterior pdf used as a surrogate model for $\tilde{r}$. For each GP model $m$, the corresponding weight $w_t^m:={\rm Pr}(i\!=\! m|\pazocal{D}_{t})$ is obtained via Bayes' rule as
\begin{align}
   w_t^m \propto p(\pazocal{D}_{t}|i\!=\! m) {\rm Pr} (i=m) =  p(\pazocal{D}_{t}|i\!=\! m) w_0^m \label{eq:weight_update} 
\end{align}
 with $p(\pazocal{D}_{t}|i\!=\! m)$ denoting the marginal likelihood of $\pazocal{D}_t$ at slot $t$. The latter in the GP regression setting is Gaussian distributed as $p(\pazocal{D}_{t}|i\!=\! m) = \pazocal{N}(\mathbf{y}_t;\mathbf{0}_t,\mathbf{K}_t^m + (\sigma_n^m)^2\mathbf{I}_t)$ where $\mathbf{K}_t^m$ and $(\sigma_n^m)^2$ are the kernel matrix and noise variance of the $m^\text{th}$ GP model respectively  \cite{Rasmussen2006gaussian}.

With the posterior pdf ${p}(\tilde{r}(\mathbf{z})|\pazocal{D}_{t})$ and the weights $\{w_t^m\}_{m=1}^M$ at hand, the next input query point $\mathbf{z}_{t+1}$ can be obtained by first selecting a certain GP model in the ensemble as ${m}_t \sim \pazocal{CAT}(\pazocal{M},\mathbf{w}_t)$, where $\pazocal{CAT}(\pazocal{M},\mathbf{w}_t)$ is a categorical distribution that draws a value from $\pazocal{M}$ with probabilities $\mathbf{w}_t:=[w_t^1,\ldots, w_t^M]^\top$. Upon selecting the GP model $m_t$, $\mathbf{z}_{t+1}$ is obtained using the following adaptive ensemble EI-based acquisition criterion 
\begin{align}
	 \mathbf{z}_{t+1} &= \underset{\mathbf{z}\in\pazocal{Z}}{\arg\max} \; \Delta_t^{m_t}(\mathbf{z})\Phi(\frac{\Delta_t^{m_t}(\mathbf{z})}{\sigma_t^{m_t}(\mathbf{z})}) + \sigma_t^{m_t}(\mathbf{z})\phi(\frac{\Delta_t^{m_t}(\mathbf{z})}{\sigma_t^{m_t}(\mathbf{z})})  \label{eq:EGP_EI} 
\end{align}
with $\Delta_t^{m_t}(\mathbf{x}):= \mu_t^{m_t}(\mathbf{x})-\hat{\tilde{r}}_t^{\text{max}}$ and $\mu_t^{m_t}(\mathbf{x})$, $\sigma_t^{m_t}(\mathbf{x})$ denoting the posterior mean and variance of the $m^\text{th}$ GP model respectively (c.f. Eqn. \eqref{eq:mean}, \eqref{eq:variance}). Note that unlike the single GP-EI criterion in Eqn. \eqref{eq:Gaussian_EI}, the advocated AF in \eqref{eq:EGP_EI}  prudently adapts to the $m^\text{th}$ GP
model at each slot $t$ as new input-output data are processed in an online fashion, balancing well the exploitation and exploration of the search space. Fig. \ref{fig:BO_nutshell} illustrates the steps at each iteration of the BO process of the advocated EGP method with adaptive EI AF, used for the VP problem of interest (henceforth abbreviated as `EGP-VP'). 

\noindent \textbf{Remark 1.} In this paper, we consider only motion noise in the reward function $\tilde{r}(\cdot)$ for the VP setting, but the proposed approach can readlily accommodate different types of noise as well. 

\noindent \textbf{Remark 2.} The present work focuses on settings where the number of available cameras $N$ is small to show that the proposed approach can offer 3D reconstruction of good quality, even with a limited budget of available cameras. If more cameras are available, they can be readily exploited by the advocated method to improve the reconstruction quality.

\begin{figure}[t]
\centering
\includegraphics[width=0.32\textwidth]{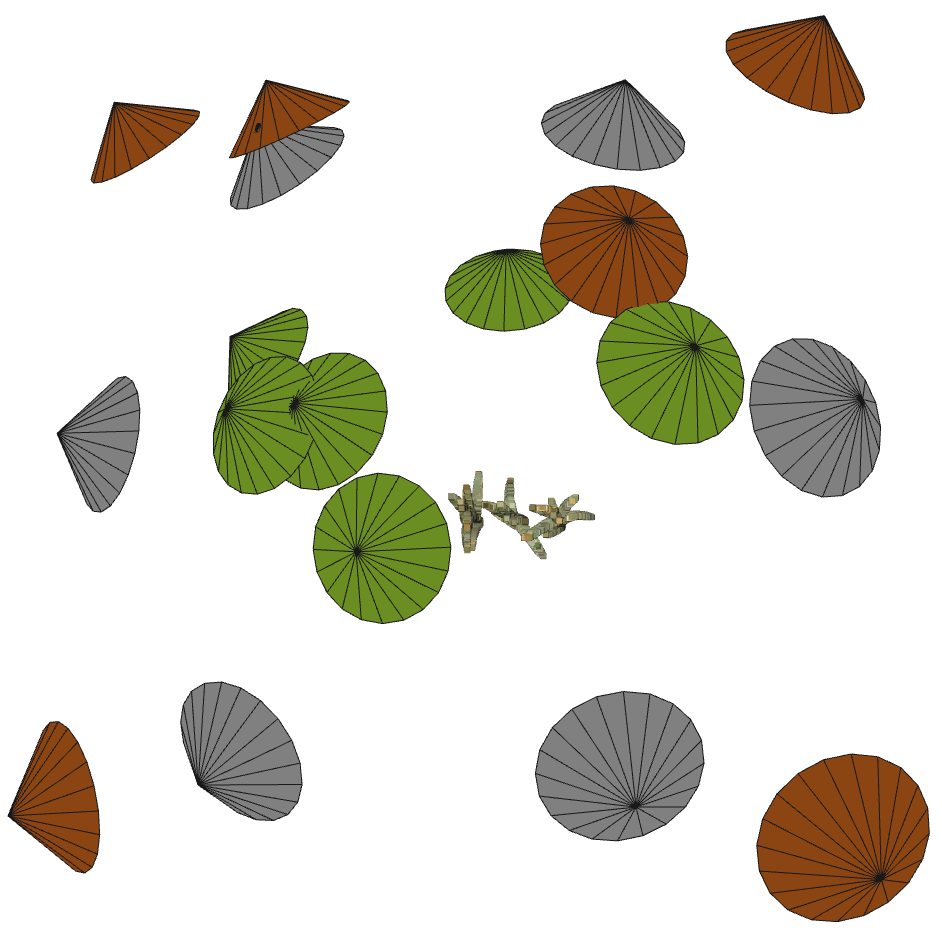}
\caption{Resulting position and orientation of all available cameras for each method, used for 3D reconstruction in the 3-plant scenario; i) gray: Circle, ii) brown: Simulated Annealing, iii) green: EGP-VP.}
\label{fig:cam_results}
\end{figure}

\begin{table*}[t]
\caption{Quantitative results of the reconstructions based on point cloud completion using the CD metric  (lower is better).}
\vspace{+0.1 cm}
\begin{center}
\begin{tabular}{|c|c|c|c|}
\hline
 &  \textbf{1-Plant (Cams=4)} & \textbf{3-Plant (Cams=6)}  & \textbf{6-Plant (Cams=6)}\\ \hline
Method     & Avg CD $\pm$ std ($\cross$ 100) & Avg CD $\pm$ std ($\cross$ 100) &  Avg CD $\pm$ std ($\cross$ 100)\\ \hline \hline
Circle                & 4.127 $\pm$ 0.194 & 2.817 $\pm$ 0.079 & 2.717 $\pm$ 0.072\\ \hline
SA & 3.844 $\pm$ 0.224 & 2.43 $\pm$ 0.192 & 3.905 $\pm$ 0.184 \\ \hline
EGP-VP (Ours)& \textbf{3.256 $\pm$ 0.245} & \textbf{1.614 $\pm$ 0.166} & \textbf{2.54 $\pm$ 0.578} \\ \hline
\end{tabular}
\label{tab:chamfer}
\end{center}
\end{table*}

\section{Experiments}


\subsection{Implementation Details}

In our experimental setup and evaluation, we used the PyRender software \cite{pyrender} for the photo-realistic scene creation and the image rendering process. We considered three different environmental scenarios with 1-plant, 3-plant (row), and 6-plant (rectangle) as illustrated in Fig. \ref{fig:env_scenarios}. The data used to create the simulated scenes were real-world corn-plant reconstructions as presented in \cite{datset}. Each plant was given a random rotation around its vertical axis and a scale within 10$\%$ of the original size to get a realistic representation of the environment. In our setup, we consider the wind as noise which is the most common source in agricultural environments that creates disturbances in the crops. 
To emulate this noise, we considered $L$ levels of the plants and for each level $l \in \{1,\ldots, L\}$ we have applied a noise value $l\times n_{\text{wind}}$ where $n_{\text{wind}} \sim \pazocal{N}(n_{\text{wind}};0, 0.002)$. The intuition is that the effect of the wind is more evident in the upper parts of the plants in real-world settings. In order to get solutions (i.e. locations and orientations of cameras) that can handle different behaviors of the wind, we considered five distinct realizations of $n_{\text{wind}}$ to create five different noisy environments. It is worth noticing that although the present work emphasizes on the wind noise, our formulation in Sec. \ref{sec:probform} can account for other types of noise as well. 

The 3D reconstruction process was conducted with COLMAP \cite{schoenberger2016sfm, schoenberger2016mvs} using the default settings. For the RGB camera parameters, we set the $FoV=\frac{\pi}{2}$, $\theta_{match} = \frac{\pi}{4}$, and the rendering resolution at $2000 \cross 1500$. 
To demonstrate the benefits of reconstructing 3D areas of interest in noisy environments with a very small camera budget, the number of cameras $N$ in our experiments was set to $N=4,6,6$ for the 1-plant, 3-plant, and 6-plant cases respectively. This mathematically means that each input vector $\mathbf{z}$ in the optimization method has dimensions $24\times 1$ for 1-plant and $36\times 1$ for the 3-plant and 6-plant cases since each camera is represented as a $6 \times 1$ vector (c.f. Sec. III). 




To optimize the noisy black-box reward function $\tilde{r}(\cdot)$, the adaptive EGP-VP framework described in Sec. \ref{sec:egpvp} was employed with the ensemble consisting of (i) a GP model with radial basis function (RBF) kernel, (ii) a GP  with an RBF kernel with automatic relevance determination (ARD), (iii) a GP  with a Matern kernel with parameter $\nu = 2.5$ and (iv) a GP  with a Matern kernel with parameter $\nu = 1.5$. For a given camera location and orientation $\mathbf{z}$ (input), the corresponding $\tilde{r}(\mathbf{z})$ (output) in the 1, 3 and 6-plant cases is the average reward function value of the five distinct noisy environments. In our experimental setting, 50 randomly selected input-output evaluation pairs $\{\mathbf{z}_\tau, \tilde{r}(\mathbf{z}_\tau)\}_{\tau=1}^{50}$ were used to train the initial parameters of the EGP-VP model and 50 iterations were considered for the BO process. Both in the BO process, where newly acquired input-output data become available online, and in the training process, the kernel parameters of each GP model in the ensemble were obtained maximizing the marginal log-likelihood via \textit{GPytorch} \cite{gardner2018gpytorch} and \textit{Botorch} \cite{balandat2020botorch} packages.


We compared our proposed EGP-VP method with two baselines; namely (i) a standard circular formation and (ii) the discrete formulation counterpart described in \cite{bacharis2022view} that uses Simulated Annealing (SA) \cite{sa} for the optimization problem. For the circular formation, we experimented with different values of the radius and the altitude and considered the best-performing one for a fair comparison, while the SA followed the same implementation as in \cite{bacharis2022view}.



\subsection{Evaluation}\label{sec:qantE}

Initially, the performance of the proposed approach in optimizing the reward function $\tilde{r}(\cdot)$ is evaluated using the simple regret (SR) metric, which at iteration $t$ is expressed as  
\begin{align}
    SR(t) := \tilde{r}(\mathbf{z}^*) - \max_{\tau\in\{1,\ldots,t\}} \tilde{r}(\mathbf{z}_\tau). \label{eq:SR}
\end{align}
Fig. \ref{fig:sr} illustrates the SR performance of all competing approaches for the (a) 1-plant, (b) 3-plant, and (c) 6-plant cases, where the optimal value $\tilde{r}(\mathbf{z}^*) = 1$ in Eqn. \eqref{eq:SR} represents the maximum reward function value. It is evident that our EGP-VP approach consistently outperforms the baselines in all cases. This showcases the significance of its benefits in balancing well the exploitation and exploration of the search space as well as the merits of adopting a continuous formulation over a discrete one as in the SA approach \cite{bacharis2022view}.

Upon obtaining the camera placement vector $\mathbf{z}^{\text{EGP-VP}}$ that maximizes the $\tilde{r}(\cdot)$ function, the reconstruction quality was quantitatively assessed using the Chamfer Distance (CD) \cite{point-cloud-utils} that is a widely used metric for point cloud completion \cite{guo2020deep, wu2021balanced}. Table \ref{tab:chamfer} reports the average CD along with the corresponding standard deviation of all approaches for five distinct noise $n_{\text{wind}}$ realizations. It can be clearly seen that our advocated EGP-VP method enjoys the lowest average CD compared to the baselines, which translates to improved average qualitative 3D reconstructions for all three environmental cases. This corroborates the effectiveness of the proposed EGP-VP approach in selecting appropriate camera placements that provide sufficient information to accurately reconstruct areas of interest, even with a limited budget of available cameras. For visualization purposes, in Figs. \ref{fig:recs} and \ref{fig:cam_results}, we demonstrate the reconstructed point clouds and the optimized camera locations for all approaches in the 3-plant case. It is evident in Fig. \ref{fig:recs} that our EGP-VP method offers the best reconstruction of the corn plants qualitatively, as can be observed by the annotated red boxes, that demonstrate substantial point cloud completeness compared to the baselines.

\section{Conclusions}

This paper focuses on the VP problem of optimally placing a given set of cameras in a 3D space, to obtain sufficient visual information and accurately reconstruct a 3D area of interest. Unlike existing approaches in the literature, this work is the first to incorporate the existing noise of the environment in the VP problem without knowledge of the analytic noise expression. To optimize the so-termed reward function that gives an estimate of the reconstruction quality and whose closed-form expression is unknown due to the embodied noise, an adaptive Bayesian optimization technique is advocated. This provides assistance in effectively optimizing the black-box reward function in a sample-efficient manner. While most existing VP approaches entail discretizing the search space, the proposed approach depends on a continuous optimization problem where any position belonging to the entire space can be explored. Numerical tests on noisy agricultural settings demonstrate the efficacy of the novel approach in accurately reconstructing 3D areas with only a small budget of available cameras. 

Future directions include theoretical and robustness analysis of the proposed method, and its assessment on real-world agricultural environments. Further environmental scenarios can additionally be considered, with applications to forestry and fire monitoring or urban surveillance.

\section{Acknowledgements}
\balance
This work is supported by the Minnesota Robotics Institute (MnRI) and the National Science Foundation through grants \#CNS-1439728,  \#CNS-1531330, and \#CNS-1939033. USDA/NIFA has also supported this work through the grants 2020-67021-30755 and 2023-67021-39829. The work of Konstantinos D. Polyzos was also supported by the Onassis Foundation Scholarship.


\typeout{}

\bibliographystyle{IEEEtran}
\bibliography{bib.bib}

\end{document}